# Aiming to Minimize Alcohol-Impaired Road Fatalities: Utilizing Fairness-Aware and Domain Knowledge-Infused Artificial Intelligence

Tejas Venkateswaran
*Computing Sciences*
*University of Hartford*
West Hartford, USA
venkatesw@hartford.edu

Sheikh Rabiul Islam
*Computer Science*
*Rutgers University - Camden*
Camden, USA
sheikh.islam@rutgers.edu

Md Golam Moula Mehedi Hasan
*Computer Science*
*Iona University*
New Rochelle, USA
mmehedihasan@iona.edu

Mohiuddin Ahmed
*Computing and Security*
*Edith Cowan University*
Joondalup, Australia
mohiuddin.ahmed@ecu.edu.au

*Abstract*—Approximately 30% of all traffic fatalities in the United States are attributed to alcohol-impaired driving. This means that, despite stringent laws against this offense in every state, the frequency of drunk driving accidents is alarming, resulting in approximately one person being killed every 45 minutes. The process of charging individuals with Driving Under the Influence (DUI) is intricate and can sometimes be subjective, involving multiple stages such as observing the vehicle in motion, interacting with the driver, and conducting Standardized Field Sobriety Tests (SFSTs). Biases have been observed through racial profiling, leading to some groups and geographical areas facing fewer DUI tests, resulting in many actual DUI incidents going undetected, ultimately leading to a higher number of fatalities. To tackle this issue, our research introduces an Artificial Intelligence-based predictor that is both fairness-aware and incorporates domain knowledge to analyze DUI-related fatalities in different geographic locations. Through this model, we gain intriguing insights into the interplay between various demographic groups, including age, race, and income. By utilizing the provided information to allocate policing resources in a more equitable and efficient manner, there is potential to reduce DUI-related fatalities and have a significant impact on road safety.

*Index Terms*—component, formatting, style, styling, insert

## I. INTRODUCTION

The National Highway Traffic Safety Administration (NHTSA) annually compiles data concerning the frequency of motor vehicle accidents instigated by drivers operating under the influence of alcohol, denoted by a Blood Alcohol Content (BAC) of 0.08 or higher at the time of collision. The compiled data indicate that alcohol-impaired driving contributes to approximately 30% of total vehicular fatalities in the United States. This figure translates to an approximate occurrence of one fatality every 45 minutes in the year 2020 due to alcohol-related crashes, even though legislative provisions in every state categorize such actions as severe transgressions. In light of these concerns, it becomes imperative to emphasize the pivotal role of more efficacious DUI testing methodologies. It is worth noting, however, that the precise interpretation of "efficacious" remains contingent on subjective interpretation.

The process of prosecuting individuals for Driving Under the Influence (DUI) is a complex procedure that occasionally involves subjective determinations. This process encompasses various stages, including the observation of the vehicle in motion, engagement with the driver, and administration of Standardized Field Sobriety Tests (SFSTs). Moreover, it is pertinent to acknowledge the existence of biases within predictive policing systems that prognosticate future criminal activities [1]–[3].

[4] assert that residents residing in neighborhoods characterized by a higher proportion of black inhabitants are more susceptible to experiencing DUI arrests, a phenomenon potentially attributable to the disproportionately elevated concentration of alcohol vending establishments and associated advertising within such localities. Similarly, according to the findings presented by [5], the rates of alcohol consumption, as well as the frequency of problematic drinking behaviors, appear more pronounced in people who are white when contrasted with other minority groups. As per the findings of [6], addressing white youth is a crucial aspect of interventions aimed at mitigating alcohol and other substance usage. The study by [5] also underscores the correlation between heightened risk factors and specific characteristics, including male gender, youthful age, elevated income level, non-cohabitation with a partner, and membership within a relatively small household.

The objective of this study is to establish the potential for reducing the frequency of fatal traffic accidents involving alcohol-impaired drivers by implementing more effective DUI testing procedures. Law enforcement personnel are mandated to adhere to The Civil Rights Act of 1964, which bars all forms of bias based on age, race, or gender, as stipulated by the Equal Protection Clause of the 14th Amendment within the U.S. Constitution. We present an approach that incorporates specialized domain knowledge to address the overall reduction of traffic fatalities perpetrated by alcohol-impaired drivers while also maintaining considerations of fairness. Given the pivotal roles that factors such as race, age, and economic circumstances play in both alcohol consumption patterns and

DUI arrests, we introduce county-wise variables, including the population percentage of non-Hispanic white individuals (the majority group), the proportion of individuals aged 65 and above, and the per capita income as our domain knowledge attributes. This infusion of domain knowledge enriches the dataset, offering a more comprehensive perspective on the matter at hand. Domain knowledge, characterized as a highly specialized form of expertise tailored to a particular field, discipline, or undertaking, significantly enhances our comprehension of the subject matter under scrutiny. Through a systematic examination, we evaluate the outcomes resulting from the incremental inclusion of these individual domain knowledge and their cumulative effects. We also applied a bias detection technique to detect the possibility of hidden bias in predicting total fatalities in an area. Subsequently, we employ a bias-mitigation approach based on reweighting to assign equitable weights to individual data points. This step subsequently corrects predictions with improved fairness and concurrently alleviates the identified biases.

Our findings elucidate the pivotal significance of specific domain knowledge variables, including the proportion of non-Hispanic white residents, the percentage of individuals aged 65 and older, and the per capita income. These variables are instrumental in comprehending the intricate interplay of factors contributing to alcohol-related fatalities. We would like to mention that we need to be mindful about including features in the prediction process that are directly related to protected attributes (e.g., race), as this might be conflicting. The inclusion of non-Hispanic whites as a domain of knowledge is solely for potential bias detection and mitigation purposes. By integrating a fairness-aware approach informed by this domain knowledge, we have successfully pinpointed regions necessitating heightened attention to curtail the overall incidence of DUI-related fatalities effectively. The outcomes of our analysis underscore the critical role of directing increased DUI intervention efforts towards locales characterized by a higher proportion of non-Hispanic white inhabitants, as evidenced by both the Theil index and balanced accuracy metrics. Furthermore, assessing disparate impact reveals the significance of considering per capita income, with areas characterized by very low or very high-income levels demonstrating increased susceptibility to DUI-related fatalities.

In the Background section, pertinent literature was examined. Subsequently, the Experiments and Results section starts with a succinct overview of the proposed architecture and dataset, and then delves into the experimental procedures, accompanied by an in-depth analysis of the outcomes.

## II. Background

Extensive research has been conducted concerning traffic accidents from intoxicated driving within society. This work encompasses interconnected themes, including factors influencing recidivism rates nationwide and within diverse racial and ethnic groups. Likewise, substantial effort has been directed toward developing strategies for mitigating these incidents, such as DUI intervention programs aimed at diminishing racial and ethnic disparities and governing the sale of alcoholic substances, among others. DUI offenses carry the potential consequence of compelling offenders to participate in mandatory treatment programs by state laws and the gravity of the offense. These programs are designed to curb recidivism through a multifaceted approach, incorporating strategies like random alcohol testing, restricting access to alcohol, fostering extended social support networks, and deploying psychotherapeutic techniques encompassing individual, group, and cognitive behavioral therapies. An exemplar of such a theory-based prevention initiative is Prime For Life, which seeks to heighten risk awareness and stimulate intrinsic motivation for change [7].

Nevertheless, the efficacy of these interventions is not guaranteed, as several external factors have been identified as correlated with elevated post-treatment DUI arrest rates. Notably, the characteristics of the neighborhood play a role, with individuals residing in communities characterized by a higher proportion of black residents being more susceptible to post-treatment DUI arrests. This phenomenon is potentially attributed to the disproportionate concentration of alcohol vendors and advertising within these localities [4]. DUI recidivism has also been linked to a constellation of problems, including aggression, negative affect, drug problem severity, and juvenile and childhood delinquencies [8]. Childhood delinquency suggests an increased tendency to disregard established rules and laws, analogous to the role of criminal activity [8]. In today's era, repeat DUI offenders are likely to suffer from coexisting drug-related problems in addition to their alcohol addictions [7].

The complexities of drinking-related issues are underpinned by various social structural, and mental health factors, exhibiting notable variations when assessed across diverse racial and ethnic population cohorts. Broadly recognized risk factors encompass male gender, young age, relatively elevated income, lack of cohabitation with a partner, and membership in smaller family units [5]. In their investigation, [5] conducted a study scrutinizing the interplay between race/ethnicity and myriad independent variables concerning alcohol consumption quantities, with whites serving as the benchmark group. Their findings underscored that the correlation between male gender and alcohol consumption was more pronounced for Hispanics than Whites. Furthermore, they noted that access to medical care was associated with a reduced frequency of drinking-related problems among Whites compared to Blacks and Hispanics.

Police are supposed to abide by The Civil Rights Act of 1964 which prohibits any discrimination based on age, race, or gender under the Equal Protection Clause of the 14th Amendment to the U.S. Constitution. Despite this however, social stigmas continue to be prevalent thus promoting the widely held notion that racial biases exist in law enforcement with police agencies being accused of this when performing public safety duties [9]. Technological progress has facilitated the implementation of automated detection systems, significantly bolstering the legal system's capacity to vigilantly monitor offenders' behaviors and prevent them from operating vehicles. Among these advancements is an emerging technology that employs ankle bracelets equipped with accelerometers to detect driving-related foot movements, providing a tangible measure of vehicle operation [10]. Additionally, the Driver Alcohol-Detection System for Safety program, jointly funded by the NHTSA and motor vehicle manufacturers, pursues the ambitious goal proposed half a century ago: to create "Cars that Drunks Can't Drive." One of the program's monitors passively identifies the presence of alcohol in a driver's breath, exemplifying the clear potential of these advancements to markedly enhance traffic safety and consequently curtail alcohol-impaired driving fatalities [10].

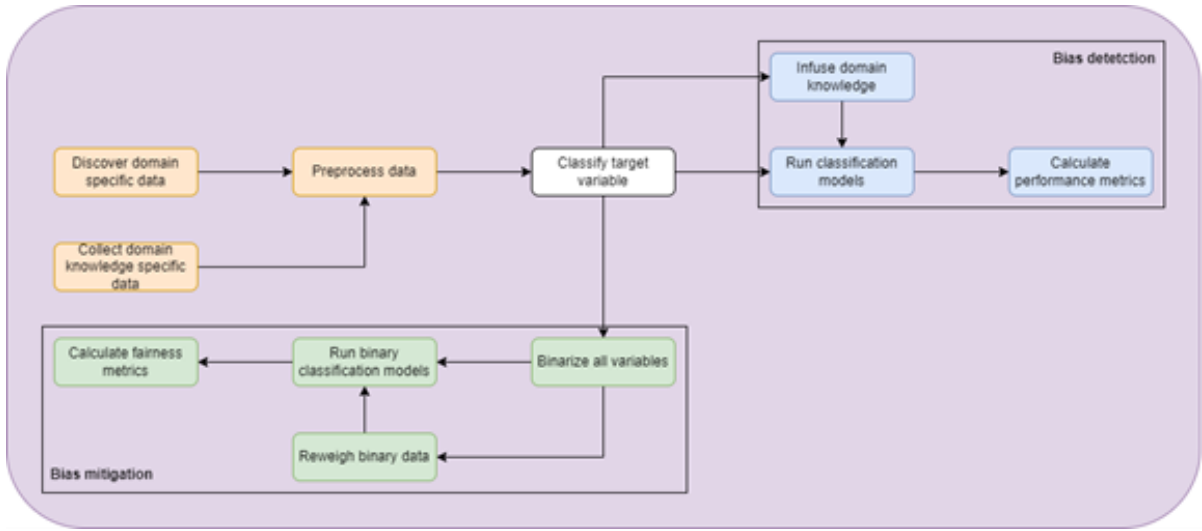

Fig. 1: Overview of the project structure. A flowchart representing the process of development for this thesis. The process starts by collecting domain-specific data, then moves to bias detection and bias mitigation. Domain knowledge is incorporated in bias detection, and IBM AIF360 is used for bias mitigation.

Notwithstanding the progress made, there remains an avenue for improvements for automated AI-based decision systems that consider critical factors such as the interplay between age, race, and income within specific geographic areas. Such considerations are pivotal in determining the precise requirements for DUI enforcement using automated systems. The existing work on age, race, and income is mainly on a small control group limited to a geographic region. Our work addresses these gaps to some extent and contributes to the ongoing efforts in this field.

## III. Proposed Architecture

Figure 1 shows an overview of the proposed architecture. The first step of this study is to collect and prepare data relevant to the project. We collected, processed, and merged data from different sources. Then useful domain knowledge is explored and collected. This curated dataset then needs to be preprocessed and the target variable, Alcohol-impaired driving death needs to be classified. Then, bias can be detected through various classification models. The final step is to find ways to mitigate bias which for this study included a reweighting-based bias mitigation model.

## IV. Experiments and Results

The first step of this study is to collect and prepare data relevant to this research. We collected, processed, and merged data from different sources. Then, useful domain knowledge is explored and collected. This curated dataset is then preprocessed so that the target variable, Alcohol-impaired driving deaths, can be classified. After that, bias is detected through various classification models. The final step is to find ways to mitigate bias, which included a reweighting-based bias mitigation model for this study.

### A. Dataset

The primary dataset is obtained from multiple data tables for the year 2022 from the "County Health Rankings and Roadmaps" website which is a program run by the University of Wisconsin Population Health Institute [11]. It consists of a total of 21 different attribute columns pertaining to 3107 rows of state counties across the contiguous United States thus excluding Alaska, the District of Columbia, Hawaii, and Puerto Rico. The percentage of alcohol-impaired driving deaths is being used as the dependent variable while the remaining 20 are our explanatory variables: Adult Obesity, Food Environment Index, Physically Inactive Adults, Access to Exercise Opportunities, Food Insecure, Limited Access to Healthy Foods, Insufficient Sleep, Uninsured Adults, Some College Degrees, School Segregation Index, Unemployment, Median Household Income, Children Single-Parent Households, Annual Average Violent Crimes, Severe Housing Problems, Drive Alone to Work, Drive Alone Long Commutes, Homeowners, Broadband Internet Access, and High School

Completion. We added three new columns (per capita income, percentage of population age 65+, no-Hispanic white population percentage) for domain knowledge. The domain knowledge is collected from a different data source [12].

### B. Data Preprocessing

The dataset without domain knowledge is concatenated with the domain knowledge columns to begin the data preprocessing. Some rows with incomplete data were removed. The problem is then converted into a binary classification problem by converting alcohol-impaired driving death% into "Risk Level" where counties' (individual records) were labeled as risk level 0 = "Low Risk" or 1 = "High" according to the number of traffic fatalities perpetrated by alcohol-impaired drivers. Counties with alcohol-impaired driving deaths of less than 21%, equal to the 30th percentile of the distribution were classified as "Low" while those at or above this value were classified as "High". The basis for choosing this cutoff percentile was that it gave the most optimal results when evaluating overall classification metric scores of models using different thresholds through trial and error. The finalized dataset is released for replication of experiments and for extending this research [13][1].

### C. Exploratory Analysis

We start with some exploratory data analysis to get a good insight into the dataset. Here are some relevant findings from the Pearson correlation analysis that gives a score between -1 and +1 where a) a score between 0 and 1 refers to a positive correlation where if one variable changes the other variable changes in the same direction, b) a score in between 0 and -1 refers a negative correlation where if one variable changes the other variable changes in the opposite direction, and c) a score of 0 refers that there is no relationship between the variables.

- Physically inactive adult is negatively correlated (-.15) with alcohol-Impaired driving death.
- Uninsured adult is negatively correlated (-.15) with alcohol-Impaired driving death.
- High School Completion % ( Adult ≥ 25) is positively correlated (.13) with alcohol-Impaired driving death.
- White non-hispanic is positively correlated (.47) with High School Completion % ( Adult ≥ 25)
- White non-hispanic is negatively correlated (-.48) with Uninsured adult
- White non-hispanic is negatively correlated (-.27) with Physically inactive adult

So far, there is no strong linear correlation (closer to 1) with the dependent variable alcohol-impaired driving death is visible. So, we investigated further and zoomed into the added three pieces of domain knowledge (See Figure 2). While they also do not exhibit any significant linear correlation with Alcohol-Impaired Driving death, Non-Hispanic White, Per Capita Income, and Age Over 65% all were slightly positively correlated with Alcohol-Impaired Driving death. In addition to that, Non-Hispanic White is highly correlated (.36) with age over 65+ and moderately correlated (.15) with per capita income. These findings interest us to dig further and explore the dynamics. It is crucial to note that the Pearson correlation does not capture any non-linear relationship that might exist.

[1]https://github.com/tejasv378/BigData-2023-DUI

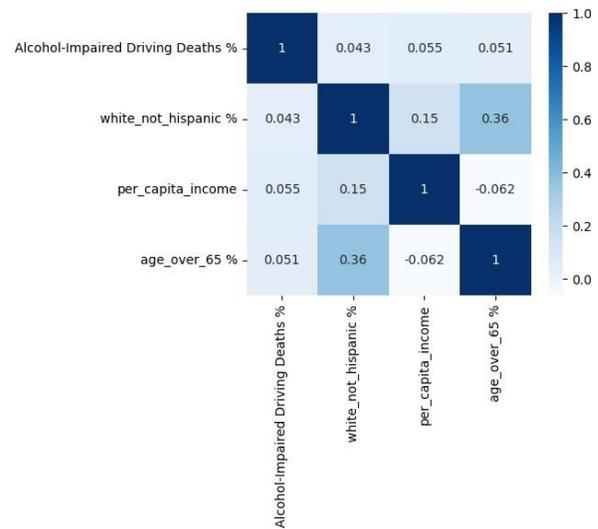

Fig. 2: Correlation analysis among domain knowledge and dependent variable.

All three charts (Figures 3a, 3b, and 3c) suggest areas with more white non-Hispanic population percentage are at higher risk (skewed towards the right) of DUI death. Figures 3a and 3b show there is the biggest dispersion (clear distinction) between low-risk and high-risk. However, areas with low non-Hispanic white (less than 20 percent) are comparatively less risky and dispersion is not clearly discernable. Figure 3c, the bivariate distributions show high density on the right middle side which means that areas with greater than 60% white non-Hispanic are at greater risk.

Regions characterized by a per capita income within the range of 25,000 to 35,000 exhibited the highest frequency, coupled with significant dispersion in the graph (Figures 4a and 4b), indicating an elevated DUI risk. This observation finds additional support in Figure 4c, which highlights a concentrated and intense zone within the specified income bracket. Correspondingly, affluent localities boasting incomes exceeding 70,000 also demonstrate a degree of susceptibility to DUI incidents. In contrast, areas featuring income ranges spanning 55,000 to 60,000 tend to exhibit a relatively neutral profile.

Areas with around 15-20% of people 65+ were more frequent and risk level dispersion was high (Figures 5a and 5b). In areas where more than 40% of people are 65+ the level of dispersion is minimal. There is a dense cluster in areas with only 20-30% of people 65+ (see Figure 5c). This suggests areas with fewer older people are riskier.

In summary, the exploratory analysis reveals that areas with more than 60% non-Hispanic white, 20k-30k income, and 20-

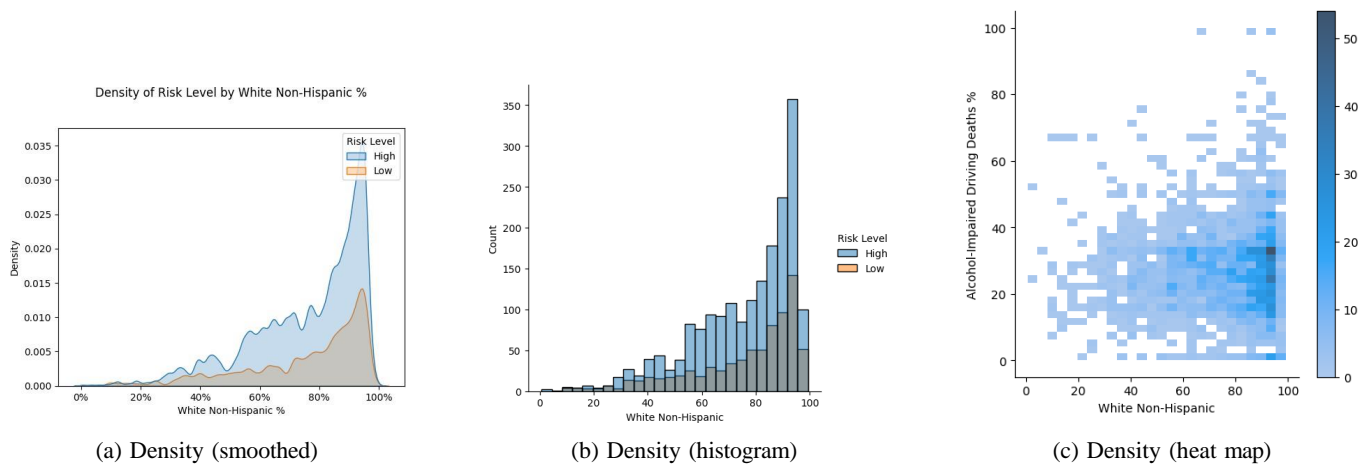

(a) Density (smoothed)     (b) Density (histogram)     (c) Density (heat map)

Fig. 3: (a) Density (smoothed) of risk level by white non-Hispanic population percentage, (b) Density (histogram) of risk level by white non-Hispanic population percentage, and (c) Density (heat map) of risk level by white non-Hispanic population percentage.

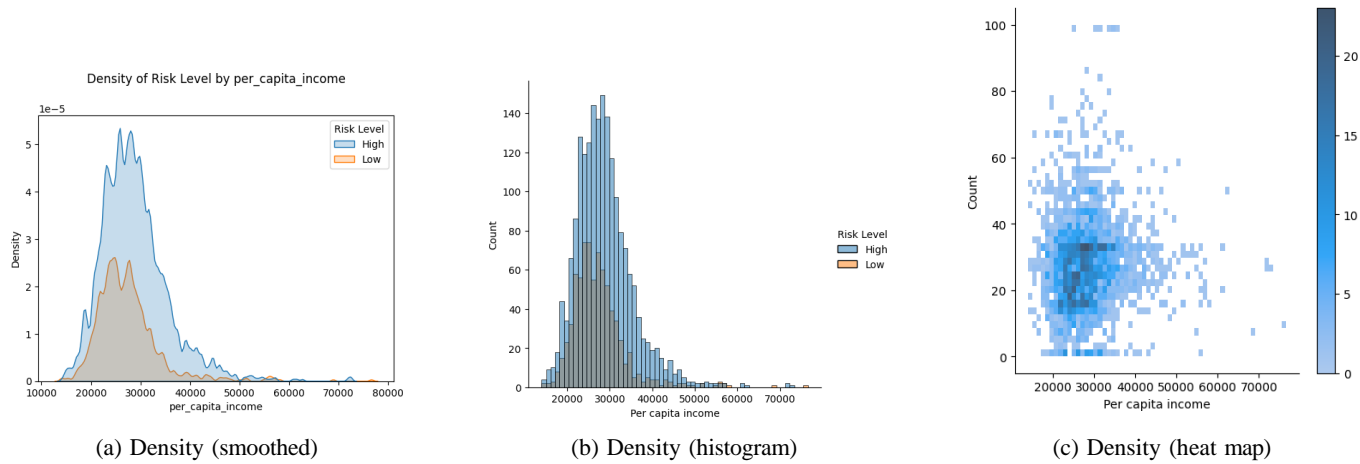

(a) Density (smoothed)     (b) Density (histogram)     (c) Density (heat map)

Fig. 4: (a) Density (smoothed) of risk level by per capita income, (b) Density (histogram) of risk level by per capita income, and (c) Density (heat map) of risk level by per capita income.

30% people 65+ are at more DUI risk. Once again, this is based on the linear relationship of variables and it omits non-linear or hidden relationships among variables.

### D. Model Development

We use Python programming language to develop machine learning models on the dataset. Jupyter notebook for created models is available here for better interpretation and replication of results [13][2]. We first investigate the classification performance of supervised algorithms in predicting the risk level (alcohol-impaired driving fatalities) of individual counties. We apply five popular supervised classification models using a 70/30 training/test split. We run Logistic Regression, Random Forests, Decision Trees, and K-Nearest Neighbor, and Support Vector Machine algorithms with and without domain knowledge using the scikit-learn package. All algorithms demonstrate similar performance in predicting the risk level of individual counties in terms of performance metrics accuracy, precision, recall, and f1-score on data before infusion of domain knowledge (see Figure 6). We use the Logistic Regression algorithm for further analysis of fairness and domain knowledge inclusion as it performs one of the top-performing and well-known algorithms.

[2] https://github.com/tejasv378/BigData-2023-DUI

Then, we investigate whether domain-knowledge inclusion adversely affects the classification performance of the Logistic Regression algorithm compared to the absence of domain knowledge. Overall, in terms of accuracy, precision, Recall, and F1, Logistic Regression Algorithm demonstrates similar results with domain knowledge compared to without domain knowledge counterpart. Figure 7 exhibits a comparison of with and without domain knowledge results for Logistic Regression demonstrating negligible to no sacrifice in performance after

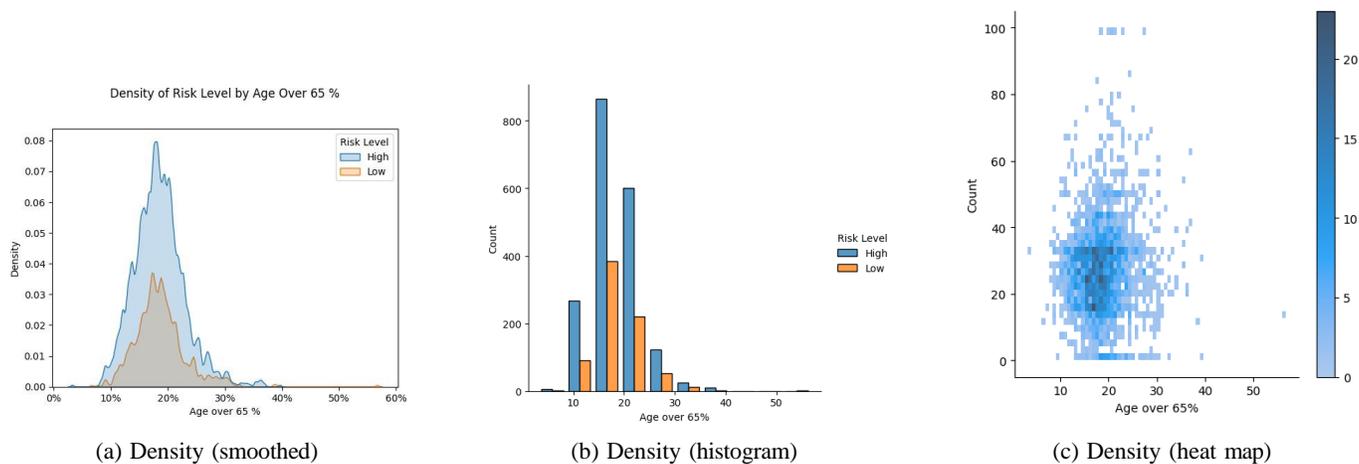

(a) Density (smoothed)   (b) Density (histogram)   (c) Density (heat map)

Fig. 5: (a) Density (smoothed) of risk level age over 65 population percentage
, (b) Density (histogram) of risk level age over 65 population percentage, and (c) Density (heat map) of risk level age over 65 population percentage.

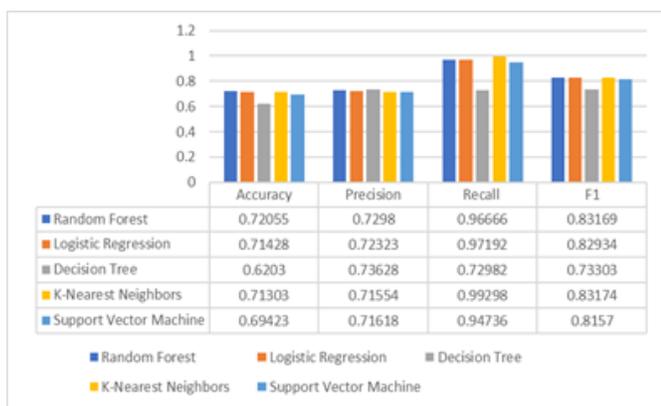

Fig. 6: Performance of different algorithms without domain knowledge.

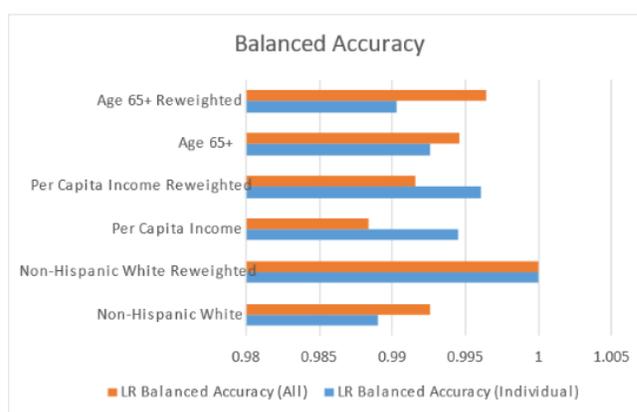

Fig. 8: Bias mitigation - balanced accuracy

### E. Bias Detection and Mitigation

We use IBM AI Fairness 360 toolkit[3] to detect and mitigate fairness-related issues. Columns containing domain knowledge are binarized based on the threshold determined from the data distribution as this is a requirement for the fairness evaluation tool. For instance, counties that had a non-Hispanic white population percentage of 59.3 or greater, with 59.3 being the national average as of 2022, were classified as "1" (having a high percentage of non-Hispanic white residents), while those having anything lower than this were classified as "0". The goal of this is to identify potential biases associated with race given that models are often susceptible to unwanted bias from protected attributes in the data. For our study, we specifically try to identify whether the algorithm might have an increased tendency to classify counties originally labeled as '1' (with a high non-Hispanic white population percentage) to label '0' (having a low-risk level for alcohol-impaired driving fatalities)

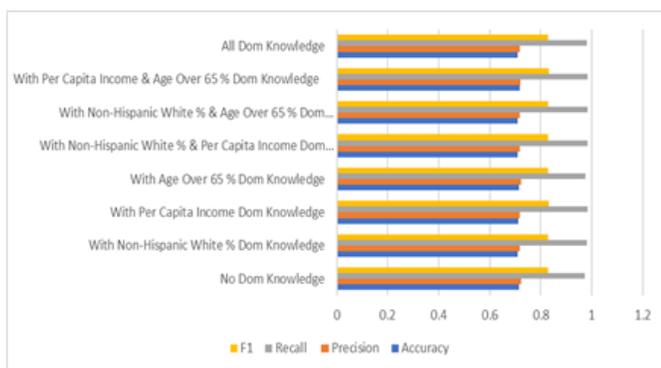

Fig. 7: Performance comparison of Logistic Regression algorithm with different combinations of domain knowledge.

infusion of domain knowledge.

[3]https://www.ibm.com/opensource/open/projects/ai-fairness-360/

during the prediction process.

Fairness has many definitions and fairness metrics accordingly. We tested four popular fairness metrics: Balanced Accuracy, Statistical Parity Difference, Disparate Impact, and Theil Index. A brief definition for each of these metrics, their fairness range, and their evaluation are discussed next. The reweighting technique is also used to help mitigate bias where AIF360 toolkit assigns weights to training set tuples instead of to class labels [14]. This is beneficial because it allows for the analysis of how protected attributes (domain knowledge columns) play a role in the development of algorithmic bias in the scope of predicting risky or non-risky counties. In other words, if the fairness is improved in the reweighted settings the domain knowledge is helping to improve fairness (i.e., reducing bias).

**Balanced Accuracy**: Balanced accuracy is the mean of the true positive rate and true negative rate of the binary classifier. The true positive rate is the proportion of positive instances correctly classified as positive and the true negative rate is the proportion of negative instances correctly classified as negative. The best score for balanced accuracy is 1 and the worst score is 0.

The best possible balanced accuracy of 1.00 was seen for the addition of non-Hispanic white domain knowledge both individually and combinedly after reweighting (see Figure 8). In all cases, balanced accuracy was improved in the combined application of all three domain knowledge. Because non-Hispanic white individually, after reweighting, can enhance balanced accuracy to the maximum, the non-Hispanic population percentage should be considered to determine DUI enforcement needs to reduce overall fatalities. This is also supported by some literature that says Whites were seen to have higher rates for both quantity of alcohol consumption and frequency of problem drinking as opposed to Blacks, Hispanics, and Asians [15]–[17].

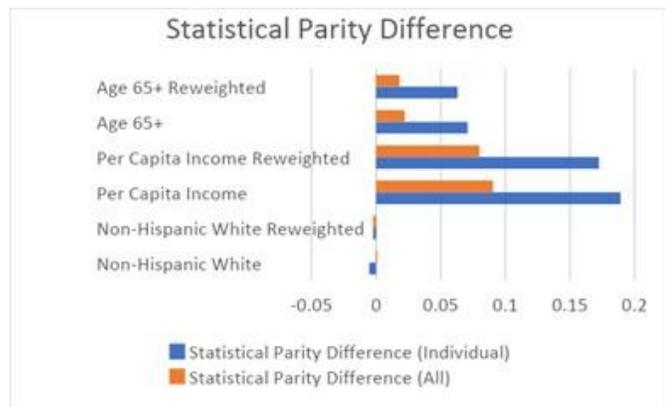

Fig. 9: Bias mitigation - statistical parity difference

**Statistical Parity Difference:** Statistical Parity Difference is the difference in probabilities of favorable outcomes for the privileged group to probabilities of favorable outcomes for the unprivileged group. It is equal to 0 when fair and the fair range of statistical parity is between -.1 to 0.1. Per Capita Income was out of fair range ($> .18$) (see Figure 9) both before and after the bias mitigation technique when added individually (blue line) but when added with the other two domain knowledge (Age 65+, Non-Hispanic White) (orange line), the statistical parity reduced to .090 (before reweighting) and .079 (after reweighting) reflecting to be within fair range (-.1 to 0.1.). Separate addition of Age 65+ and Non-Hispanic White doesn't improve or degrade the statistical parity difference. Therefore, there was a fairness issue with the per capita income of an area which was mitigated in the reweighted methods when all three pieces of domain knowledge were present. To reduce overall DUI fatalities all three pieces of domain knowledge are important to determine the enforcement needs.

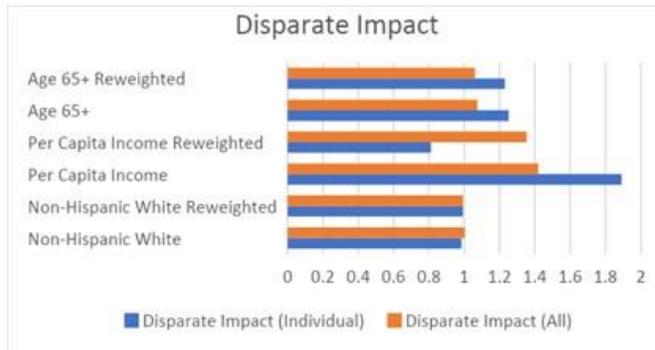

Fig. 10: Bias mitigation - disparate impact

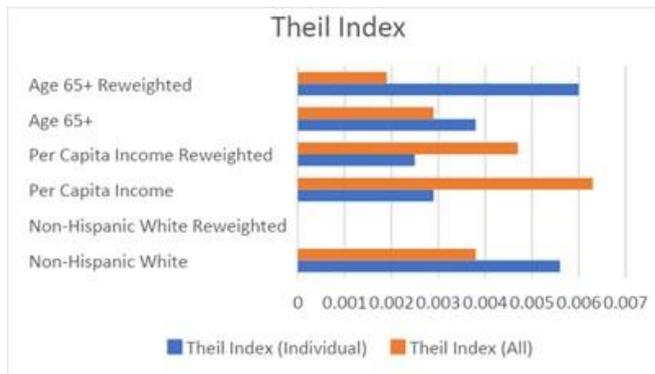

Fig. 11: Bias mitigation -Theil Index

**Disparate Impact:** The disparate impact is fair when it is equal to 1 and represents the ratio of a favorable outcome for the unprivileged group over the favorable outcome of the privileged group. The fair range is between 0.8 to 1.25. Before reweighting, per capita income is way out of the fair range individually (1.89) and combined (1.42) (see Figure 10). After reweighting, the metric becomes .81 (within fair range) for the exclusive addition of Per capita Income as domain knowledge and 1.35 (slightly above the upper bound of fair range 1.25) combined. This suggests per capita income of an area is an important indicator for DUI fatalities in that area. From the exploratory analysis section, we have seen that very low and very high-income areas face more fatalities. To be specific, areas with 25-35k were most counties in and the dispersion was high (riskier). Rich areas 70k+ were a bit risky, and 55-60k area was kind of neutral - kind of equal high and low.

**Theil Index:** The Theil index is also fair when it is equal to 0 and is a measure of the inequality in benefit allocation for individuals. For all three-domain knowledge, Theil Index reduced in the reweighted stage with the most significant progress for non-Hispanic white where the score becomes zero (no bias) after reweighting (see Figure 11). The Theil index is primarily used to measure economic inequality and other economic phenomena, sometimes, it is also used to measure racial segregation. This clearly suggests areas with more non-Hispanic whites were considered to be less risky before bias mitigation intervention and needs more intervention to reduce overall future fatalities. These findings align with some of the exploratory findings - Houseowner and Per capita Income is positively correlated (.41 and .15 accordingly) with white non-Hispanic, thus, after reweighting, this index becomes zero. Reweighting assigns weight to tuples instead of class level so that protected attributes are similarly affected. In other words, more DUI charges in areas with more non-Hispanic whites can reduce DUI fatalities in that area.

Overall, we have seen that more DUI intervention is needed in areas with more non-Hispanic white as evidenced by Theil Index and balanced accuracy. Disparate impact suggests per capita income needs to be considered as very low-income and very high-income areas are more prone to DUI fatalities. Statistical parity difference suggests all three pieces of domain knowledge are important.

## V. CONCLUSION

In this research endeavor, we have developed an AI-based prediction system that incorporates domain knowledge and fairness considerations to predict alcohol-related fatalities across different demographic areas. Our primary goal was to conduct a comparative analysis and gain insights into resource requirements for minimizing these fatalities. Overall, our findings indicate that specific domain knowledge, such as the percentage of non-Hispanic white population, 65+ population percentage, and per capita income, plays a crucial role in understanding the interplay of factors influencing alcohol-related fatalities. By incorporating fairness analysis using this domain knowledge, we identified areas that require greater focus to reduce overall DUI-related fatalities. The results highlight the importance of implementing more DUI intervention in areas with a higher percentage of the non-Hispanic white population, as indicated by the Theil index and balanced accuracy. Furthermore, the disparate impact analysis suggests that per capita income should be considered, with very low-income and very high-income areas being more susceptible to DUI fatalities. We acknowledge that the selection of domain knowledge was based on heuristic and trial-and-error methods, which provided valuable insights. However, moving forward, a systematic approach is necessary to identify and utilize domain knowledge effectively to potentially enhance our results. In addition, another extension of this work could be the categorization and analysis of fatal vs no-fatal DUI incidents, and make recommendations according to the severity of the incident. In addition, another extension of this work could be the categorization and analysis of fatal vs no-fatal DUI incidents, and make recommendations according to the severity of the incident.